# Coping in Crisis: Computational Modeling of Coping Styles in Digital Crisis Discourse During the 2023 Türkiye Earthquake

Şevval Çakıcı
Scakici25@ku.edu.tr
Koç University

**Abstract**

How do people cope when disaster strikes and can we detect it at scale, in real time, from what they write? This study addresses that question using over one million Turkish-language tweets posted in the aftermath of the February 6 2023, earthquake in Türkiye, which unfolded in a deeply polarized political context just months before a national election. Drawing on Lazarus and Folkman's (1984) coping theory, we develop a multi-label BERTurk classifier to detect three coping styles (problem-focused, emotion-focused, and meaning-making) across four theoretically motivated crisis phases. BERTurk achieves a macro F1 of 0.693, substantially outperforming a zero-shot mDeBERTa baseline (macro F1 = 0.324). Applied to the full corpus, the classifier reveals a clear temporal trajectory: problem-focused coping dominates the urgency phase and declines sharply, emotion-focused coping rises and stabilizes, and meaning-making increases monotonically. Anger correlates most strongly with meaning-making (Spearman r = 0.387), suggesting it functions as a mobilizing force toward blame attribution rather than practical action. These findings demonstrate that coping theory can be reliably operationalized in real-world digital crisis data and that doing so can help humanitarian organizations tailor their responses to where a population actually is.

## 1. Introduction

Every year, millions of people around the world face a moment they did not anticipate. Some lose their homes to floods, their loved ones to earthquakes or pandemics, their belongings to wildfires, their freedom to political decisions. Between 2000 and 2019 alone, 7,348 major natural disasters affected more than 4.2 billion people (UNDRR, 2020). This figure does not include the millions more displaced by conflict, political violence, or economic collapse. The numbers are devastating, but they do not tell us much about what actually happens to people in those moments. What did they think and feel? And how did they interpret what happened to them — how did they cope?

After a major crisis, traditional research generally waits for weeks. Surveys are distributed; interviews are planned. People are asked to describe what they felt, how they coped, what helped them get through it. This approach has produced decades of valuable findings. But memory is reconstructive. What someone reports in a calm room, weeks after the fact, is not always the same as what they actually experienced in the moment.

During the February 6, 2023 earthquake in Türkiye, people were already writing on social media within minutes of the first tremors. They shared coordinates of those trapped under rubble, expressed grief, asked why the buildings had fallen the way they did. Over the following month, more than two million tweets documented a society in crisis, in real time, in its own words. It was an unfiltered, timestamped record of how millions of people responded to one of the deadliest earthquakes in the country's modern history and written under actual stress.

According to coping theory (Lazarus & Folkman, 1984), individuals facing extreme stress deploy systematic psychological strategies, ranging from practical problem-solving to emotional regulation and meaning construction, to manage the demands of an overwhelming situation. The February 6 earthquake did not unfold in a political vacuum. In the days that followed, it became clear that thousands of buildings had collapsed not because of the earthquake's magnitude alone, but because of decades of inadequate construction standards and weak enforcement. Rescue operations were slow to reach many areas. Public anger grew quickly and it had a clear target: the state.

This matters for understanding what people wrote. Türkiye in early 2023 was already a deeply polarized society, three months away from a national election in which the ruling party faced its most serious electoral challenge in two decades. In this environment, the boundary between grief and political blame was unusually thin. People were mourning and furious at the same time, often in the same sentence. Expressing loss and demanding accountability were not separate acts, they were intertwined. This makes the 2023 earthquake an especially productive, and complicated, setting for studying how people cope in public: the data captures not just psychological responses to loss, but the entanglement of trauma and politics that defines collective crisis in polarized societies.

Despite decades of coping research, the theory has almost exclusively been tested in laboratory settings and self-reported surveys. These methods are useful, but they ask people to reflect on how they coped after the fact. Social media is different. When a disaster happens, people write in real time, under actual stress, without being asked. This gives us something rare: a large, naturally occurring record of how millions of people responded to the same event, at the same time, in their own words. Yet social media data is not a transparent window into psychological states. People do not simply express how they feel online. They also perform, self-censor, and respond to social norms about what is appropriate to say in public during a crisis. A tweet written under stress is still a public act, shaped by audience awareness, platform affordances, and the visibility of the crisis itself. This means that what we observe is not coping behavior as such, but its public, textual trace: a distinction that matters for how findings should be interpreted.

Yet coping theory was originally designed to describe how one person responds to stress, not observing how a whole population does. This study makes an important shift: instead of asking how an individual copes, we ask how collective coping expressions change over time across a population. We are not claiming that a tweet is a direct window into someone's psychology. But when hundreds of thousands of people write about the same disaster, patterns emerge and those patterns can tell us something real about how societies process collective trauma.

This study addresses that gap. The question is then: How do different psychological coping styles appear in digital discourse during the February 6 earthquake in Türkiye, and how are they related to blame, help-seeking, and solidarity patterns across time? To answer this question, we develop a multi-label classification pipeline using BERTurk, a Turkish BERT model, trained on 500 human-annotated tweets. We then apply the trained classifier to over one million tweets spanning a 30-day post-disaster window, enabling fine-grained temporal analysis of coping style dynamics across four theoretically motivated crisis phases.

Our study makes three contributions. First, it provides the first large-scale empirical test of coping theory in real-world digital crisis discourse, extending a predominantly lab-based literature to an organic, high-stakes setting. Second, it demonstrates the methodological feasibility of coping style detection in a non-English, low-resource context using a language-specific transformer model. Third, it offers substantive findings on how collective coping dynamics unfold across crisis phases, findings that speak to both crisis communication theory and the political sociology of disaster. Beyond academic contribution, this study points toward a practical horizon: if coping styles can be detected at scale and in real time, they can inform how humanitarian organizations, crisis communicators, and public institutions respond. A population dominated by problem focused coping signals a need for information and coordination. A population in the emotion focused phase signals a need for acknowledgment and space to grieve. A population turning toward meaning-making signals a need for institutional transparency and accountability.

## 2. Literature Review

### 2.1. Coping Theory

Lazarus and Folkman's (1984) transactional model of stress and coping remains the foundational framework for understanding how individuals respond to threatening events. The model identifies four primary coping strategies: problem-focused coping, which involves active attempts to address the source of stress; emotion-focused coping, which regulates affective responses without necessarily resolving the stressor; meaning-making, which involves constructing cognitive or spiritual frameworks to interpret the experience (Park, 2005); and avoidance, which entails cognitive or behavioral distancing from the stressor (Carver et al., 1989).

Coping theory was developed primarily within Western, individualistic psychological traditions. Its application to a society such as Türkiye, which sits closer to the collectivist end of cultural dimensions frameworks but has been described as having a mixed individualist-collectivist profile, raises questions about cross-cultural validity (Hofstede, 2001).

A critical limitation of this literature is its reliance on retrospective self-report instruments such as the Ways of Coping Questionnaire (Folkman & Lazarus, 1988). These methods capture how individuals reflect on past coping behavior rather than observing it as it unfolds. The shift to naturally occurring digital data does not simply solve this problem. It replaces one set of limitations with another. What is gained is temporal immediacy and scale. What is lost is depth: a tweet cannot tell us why someone wrote what they wrote, what they felt before or after, or how their coping evolved over days and weeks. This trade-off is not a reason to avoid digital data, but it is a reason to be precise about what claims it can and cannot support.

### 2.2. Crisis Communication and Social Media

Research in crisis communication has shown that Twitter (known as X after 2023) functions as a critical infrastructure for real-time disaster response (Coombs, 2015). Studies of major events (including the 2010 Haiti earthquake, the 2011 Fukushima disaster, and the 2016 Nice attack) have documented predictable temporal dynamics: early phases dominated by urgent information-seeking and coordination, middle phases characterized by blame attribution and accountability demands, and later phases marked by memorialization and meaning-making (Coombs, 2015; Jin, 2009; Imran et al., 2015).

Jin (2009) specifically links cognitive appraisal of emotions in crises to coping and communication strategy selection, providing a theoretical bridge between coping psychology and crisis communication. However, Jin's work is based on experimental designs with hypothetical crisis scenarios. The present study extends this line of research by testing whether the predicted patterns manifest in organic social media data from an actual mass casualty event.

### 2.3. Computational Approaches to Crisis Discourse

The automated analysis of crisis-related social media has grown substantially, with applications in humanitarian response, rumor detection, and situational awareness (Grimmer & Stewart, 2013; Imran et al., 2015). Transformer-based models have enabled high-accuracy classification of disaster tweets across multiple dimensions (Wolf et al., 2020; He et al., 2021). However, existing studies predominantly focus on English-language data, and none have systematically operationalized coping theory as a classification target. The use of non-English, culturally specific models such as BERTurk (Schweter, 2020) addresses a persistent gap in the computational treatment of low-resource crisis contexts.

Hovy and Spruit (2016) caution that NLP systems applied to high-stakes social data carry significant ethical responsibilities, particularly when the subjects are vulnerable populations in crisis. This study takes those concerns seriously by treating the corpus as evidence of collective psychological response rather than as a surveillance or behavioral prediction tool.

A growing body of work applies NLP to detect psychological states in social media text. These studies demonstrate that linguistic patterns carry meaningful psychological signal even in short, informal texts (Tausczik & Pennebaker, 2010). However, they focus predominantly on individual-level clinical outcomes rather than collective coping dynamics, and rarely engage with crisis communication theory. The present study sits at the intersection of these two traditions.

## 3. Data

### 3.1. Corpus

The primary data source is the Politus corpus, a large-scale Turkish political and social media dataset collected by Yörük et al (2024). The earthquake-relevant subset was identified using keyword filtering: #deprem, #6Şubat, #KahramanmaraşDepremi, #Kahramanmaraş, #Gaziantep, "deprem yardım", "enkaz", "AFAD", and "earthquake Turkey". Of 22.5 million tweets in the 6 February 6–March 6, 2023 collection window, 2,419,224 matched at least one keyword. After filtering for Turkish-language original tweets (excluding retweets) of at least 20 characters, the final analysis corpus comprised 1,037,663 tweets.

Each tweet in the corpus includes pre-computed Politus model scores for emotion, ideology, and welfare, as well as engagement metrics (like count, retweet count, impression count, reply count) and author demographic predictions (gender, age group, organization status). The availability of these auxiliary variables enables multivariate analysis of coping style correlates beyond the classification task itself.

### 3.2. Temporal Phase Structure

Following prior crisis communication literature (Coombs, 2015), the 30-day window was divided into four theoretically motivated phases. Phase 1 (Days 1–3, February 6–8) captures the urgency and immediate rescue coordination period. Phase 2 (Days 4–7, February 9–12) corresponds to the accountability and blame attribution phase. Phase 3 (Weeks 2–3, February 13– 20) captures the grief and solidarity period. Phase 4 (Week 4+, February 21–March 6) corresponds to the meaning-making and recovery phase. The corpus distribution across phases reflects the temporal concentration of earthquake discourse: Phase 1 contained 492,334 tweets (47.5%), Phase 2 287,087 (27.7%), Phase 3 151,375 (14.6%), and Phase 4 106,867 (10.3%).

### 3.3. Annotation

A stratified random sample of 500 tweets was drawn with 125 tweets per phase using reservoir sampling (random seed 42). The sampling procedure filtered for Turkish-language original tweets of at least 30 characters to ensure sufficient textual content for annotation.

From this 500-tweet sample, a stratified subset of 50 tweets (12–13 per phase) was independently labeled by three human annotators and GPT-4o using a multi-label scheme with four categories derived from Lazarus and Folkman (1984): problem-focused, emotion-focused, meaning-making, and avoidance. The codebook[1] defines each category with operational definitions and tweet-level examples. Gold standard labels for these 50 tweets were derived through majority vote across the three human annotators.

Inter-annotator agreement was measured using Cohen's Kappa for all pairwise combinations of the four annotators (three human + GPT-4o) on this 50-tweet validation set (n = 50). The avoidance category produced near-zero agreement across all pairs (mean κ = 0.021), reflecting both its theoretical ambiguity and its empirical rarity in the corpus (GPT-4o assigned it to only 4.6% of tweets). Following established practice for low-reliability categories (Krippendorff, 2004), avoidance was excluded from downstream classification and analysis.

[1] The annotation codebook is available in the analysis notebook at: https://github.com/sevvalckc/cssm530-coping-crisis

Agreement for the remaining three categories was: problem-focused κ = 0.766, emotion-focused κ = 0.591, meaning-making κ = 0.449, consistent with acceptable thresholds for complex multi-label annotation tasks.

For the remaining 450 tweets, GPT-4o labels were used as gold standard, given the high human-GPT agreement observed on the validation set (96%, 96%, and 86% for problem-focused, emotion-focused, and meaning-making, respectively). The total 500-tweet gold standard dataset thus comprises 50 human majority-vote labels and 450 GPT-4o labels.

## 4. Method[2]

### 4.1. Classification Model

The primary classification model is BERTurk[3] (dbmdz/bert-base-turkish-cased; Schweter, 2020), a BERT-base model pre-trained on a large Turkish corpus[4]. The model was fine-tuned for multi-label sequence classification using sigmoid activation with binary cross-entropy loss, enabling independent prediction of each coping style. Maximum sequence length was set to 128 tokens. Training used the AdamW optimizer with learning rate 2e-5 and weight decay 0.01 for five epochs, with early stopping based on macro F1[5] on the development set.

---

[2] All code, annotated data, and pre-computed results are available at: https://github.com/sevvalckc/cssm530-coping-crisis

[3] A note on methods for non-specialist readers. BERTurk is a version of BERT, a large language model pre-trained on Turkish text, that has learned statistical patterns of the Turkish language from a large corpus. Fine-tuning means we further train this model on our labeled earthquake tweets so that it learns to recognize coping-relevant language specifically. Language-specific models like BERTurk consistently outperform multilingual models (models trained on many languages at once) on single-language tasks because they develop deeper representations of that language's vocabulary, grammar, and idiomatic expressions. The zero-shot baseline we compare against, mDeBERTa, is such a multilingual model. It has never seen our labeled data and must rely on its general cross-lingual knowledge, which explains the large performance gap.

[4] BERTurk (transformers==4.40.0), mDeBERTa (transformers==4.40.0), Python 3.12, PyTorch 2.3.0, scikit-learn 1.4.2.

[5] Two metrics deserve a brief explanation. Macro F1 averages the F1 score, a balance of precision and recall, equally across all three coping categories, regardless of how often each appears in the data. This means a model that performs well on rare categories is rewarded equally to one that performs well on common ones. Hamming loss measures the proportion of individual label predictions that are wrong in a multi-label setting; lower is better.

The 500-tweet annotated dataset was split into training (70%, n = 350), development (15%, n = 75), and test (15%, n = 75) sets using stratified sampling by phase to ensure balanced phase representation across splits.

### 4.2. Zero-Shot Baseline

As a baseline, zero-shot classification was performed using mDeBERTa-v3-base fine tuned on MNLI and XNLI (MoritzLaurer/mDeBERTa-v3-base-mnli-xnli). Each coping category was expressed as a natural language hypothesis: "This tweet contains information sharing, aid coordination, or practical help" (problem-focused); "This tweet expresses grief, fear, gratitude, or emotional support" (emotion-focused); "This tweet frames the crisis through moral, religious, or political meaning" (meaning-making). Labels were assigned when hypothesis entailment scores exceeded 0.5 in multi-label mode.

### 4.3. Evaluation Metrics

Model performance was evaluated on the held-out test set using macro-averaged F1 (primary metric), per-label F1, Hamming loss, and subset accuracy. Statistical comparison between BERTurk and the zero-shot baseline was conducted using McNemar's test on per-label correct/incorrect classifications. The trained BERTurk model was then applied to the full 1,037,663-tweet corpus in batches of 64.

### 4.4. Temporal and Correlation Analysis

Coping style prevalence was computed as the proportion of tweets classified positive for each label, aggregated by day and by crisis phase. Spearman rank correlations were computed between coping style predictions and (a) engagement metrics (likes, retweets, impressions) and (b) Politus emotion scores (anger, fear, sadness, surprise, disapproval). All correlation tests were two tailed with $\alpha = 0.05$.

## 5. Results

### 5.1. Classifier Performance

Table 1 presents the performance of BERTurk and the zero-shot baseline on the held-out test set (n = 75).

**Table 1. Model Performance on Test Set**

| Metric | BERTurk | Zero-Shot | Δ | McNemar p |
|---|---|---|---|---|
| Macro F1 | 0.693 | 0.324 | +0.369 | - |
| F1 Problem-Focused | 0.755 | 0.361 | +0.394 | <.001* |
| F1 Emotion Focused | 0.704 | 0.162 | +0.542 | .143 n.s. |
| F1 Meaning-Making | 0.619 | 0.450 | +0.169 | .327 n.s. |
| Hamming Loss | 0.222 | 0.409 | -0.187 | - |
| Subset Accuracy | 0.427 | 0.147 | +0.280 | - |

*Note: * p<.001 (McNemar's exact test). N.s. = not significant.*

BERTurk substantially outperforms the zero-shot baseline across all metrics. The improvement is most pronounced for emotion-focused coping (ΔF1 = +0.542), where the zero shot model fails to capture Turkish-language emotional expressions. The McNemar test confirms that BERTurk's advantage is statistically significant for problem-focused classification ($p < .001$) but not for emotion-focused or meaning-making categories, likely reflecting the small test set size (n = 75) rather than a genuine absence of difference.

### 5.2. Temporal Coping Dynamics

Table 2 presents the mean coping rate per phase across the full corpus (n = 1,037,663). Figures 1 and 2 visualize these patterns.

**Table 2. Coping Style Distribution Across Crisis Phases**

| Phase | n | Date Range | Problem-F. | Emotion-F. | Meaning-M. | Dominant |
|---|---|---|---|---|---|---|
| Phase 1: Urgency | 492,334 | Feb 6–8 | 0.562 | 0.320 | 0.124 | Problem |
| Phase 2: Blame | 287,087 | Feb 9–12 | 0.389 | 0.439 | 0.237 | Transition |
| Phase 3: Grief | 151,375 | Feb 13–20 | 0.198 | 0.566 | 0.352 | Emotion |
| Phase 4: Meaning | 106,867 | Feb 21–Mar 6 | 0.194 | 0.543 | 0.373 | Emotion |

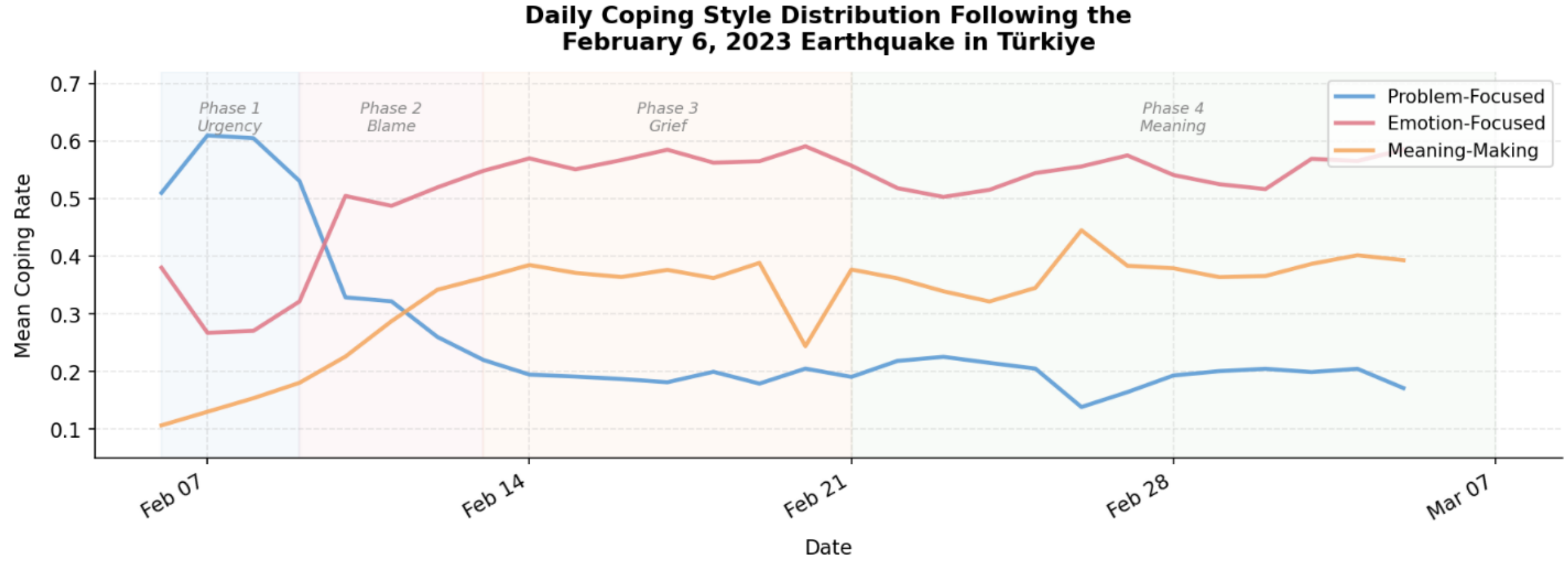

*Figure 1. Daily coping style distribution following the February 6, 2023 earthquake in Türkiye (n = 1,037,663). Shaded bands indicate crisis phases.*

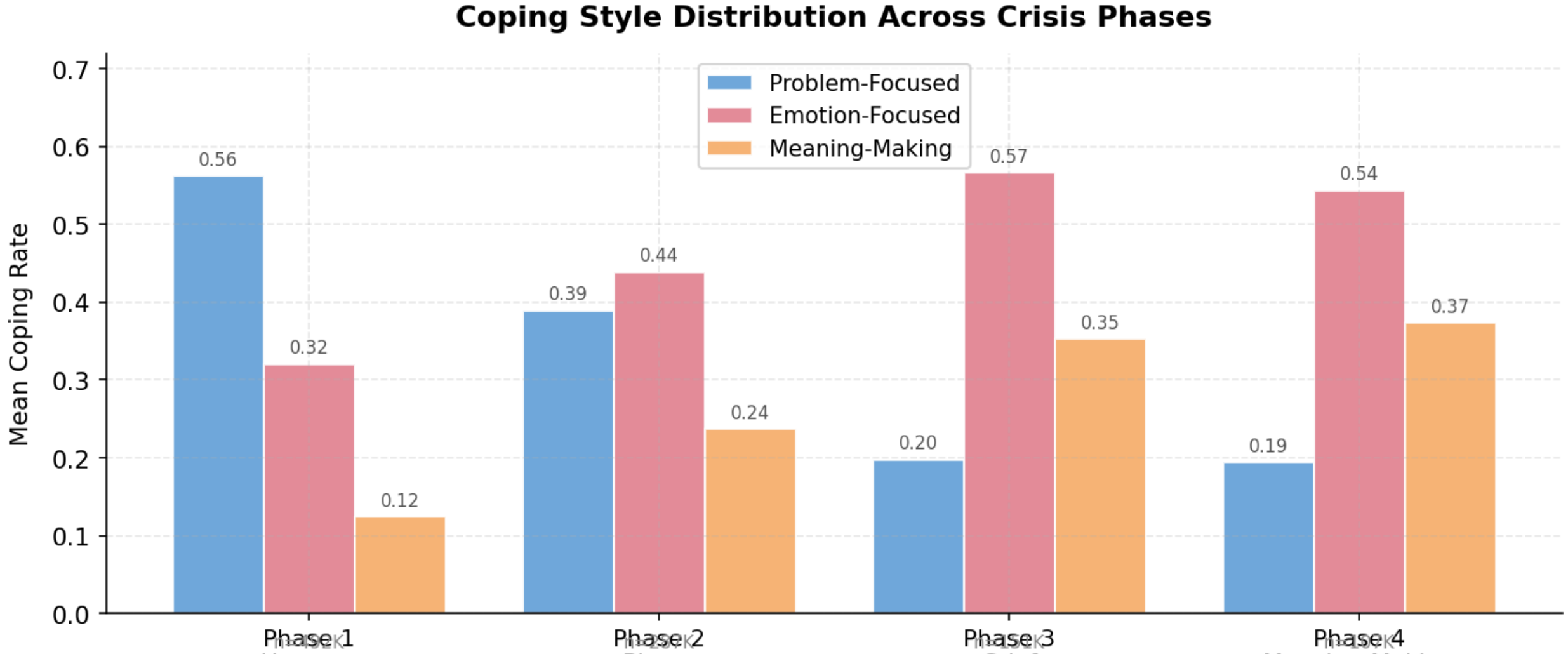

*Figure 2. Mean coping rate by crisis phase. Error bars not shown; n per phase indicated in Table 2.*

Three patterns are noteworthy. First, problem-focused coping declines monotonically from Phase 1 (0.562) to Phase 4 (0.194), a drop of 65.5%. This trajectory reflects the exhaustion of actionable rescue and coordination needs as the immediate emergency phase passes. Second, emotion-focused coping follows an inverse pattern, rising from Phase 1 (0.320) to a plateau in Phases 3 and 4 (0.566 and 0.543). A temporary dip in emotion-focused coping around February 21 likely reflects a brief resurgence of problem-focused coordination activity, coinciding with renewed public debate about shelter and aid distribution in the third week following the earthquake, before emotion-focused coping stabilizes at its plateau level in Phase 4. Third, meaning-making increases steadily across all four phases (0.124 → 0.237 → 0.352 → 0.373), consistent with the theoretical prediction that sense-making processes intensify as the acute crisis recedes and communities seek interpretive frameworks for collective loss.

### 5.3 Demographic Variation in Coping Styles

Figure 3 presents mean coping rates broken down by gender, account type, and age group across the full corpus. Several patterns are noteworthy. Female users show higher problem-focused coping rates than male users (0.510 vs. 0.388), while male users show higher emotion-focused (0.429 vs. 0.369) and meaning-making rates (0.227 vs. 0.182). Organizational accounts show markedly lower meaning-making rates than personal accounts (0.129 vs. 0.228), suggesting that institutional actors communicate in more operationally focused ways during crisis. Age group differences follow a clear gradient: younger users (≤18 and 19–29) show higher problem-focused rates, while older users (30–39 and ≥40) show higher emotion-focused and meaning-making rates, consistent with developmental differences in coping repertoires (Aldwin & Revenson, 1987; Folkman & Lazarus, 1988). These demographic patterns are exploratory and should be interpreted with caution given the limitations of automated demographic prediction in the Politus corpus.

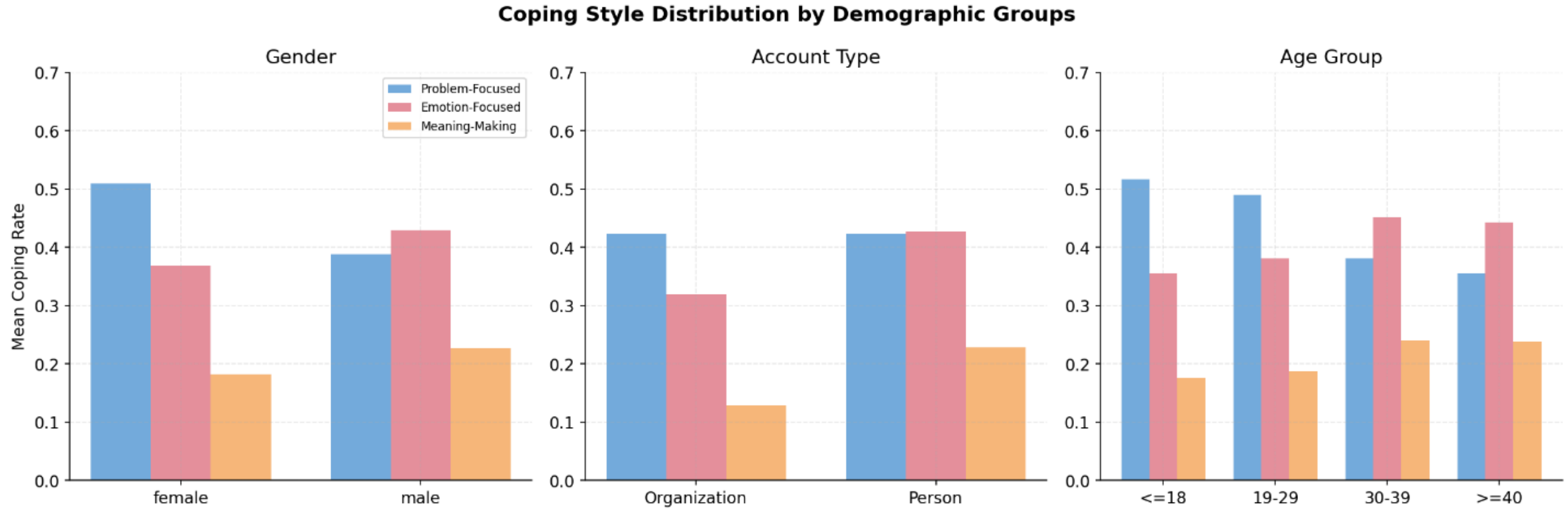

*Figure 3. Mean coping style rates by gender, account type, and age group (n = 1,037,663).*

### 5.4 Engagement Correlations

Spearman rank correlations between coping style predictions and engagement metrics reveal that problem-focused tweets show a negligible negative association with likes ($r = -0.08$) but a small positive association with retweets ($r = 0.13$), consistent with their informational character: they are shared for utility but do not generate the emotional resonance that drives like behavior. Meaning-making tweets show a small negative association with retweets ($r = -0.10$), suggesting that interpretive and political content is less likely to be amplified through sharing. Impression counts show near-zero correlations with all three coping styles.

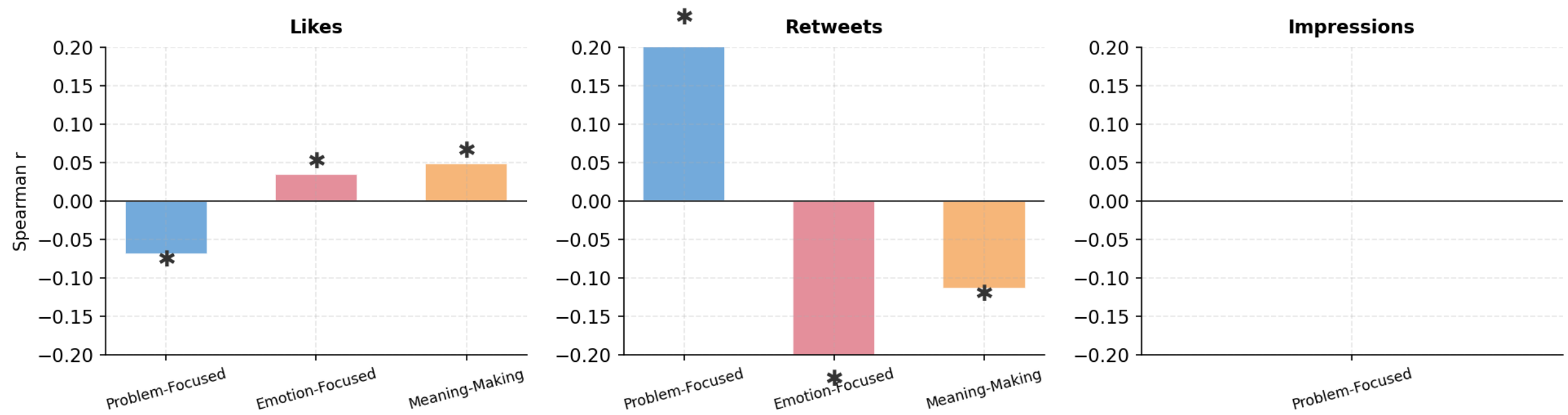

*Figure 4. Spearman correlations between coping styles and engagement metrics. *p < .05.*

### 5.5 Emotion–Coping Correlations

Given the corpus size (n > 1,000,000), all Spearman correlations reach statistical significance; interpretation therefore focuses on effect magnitude rather than significance alone. Table 3 presents Spearman correlations between Politus emotion scores and predicted coping styles. Figure 5 visualizes the full pattern.

**Table 3. Spearman Correlations: Politus Emotion Scores × Coping Styles**

| Emotion | Problem-Focused | Emotion-Focused | Meaning-Making |
|---|---|---|---|
| Anger | −0.285 * | 0.252 * | 0.387 * |
| Fear | −0.046 * | 0.053 * | 0.027 * |
| Sadness | −0.123 * | 0.184 * | 0.065 * |
| Surprise | −0.077 * | 0.066 * | 0.064 * |
| Disapproval | −0.179 * | 0.096 * | 0.221 * |

*Note: *All correlations statistically significant at p < .05 given n > 1,000,000; effect sizes are small to modest throughout.*

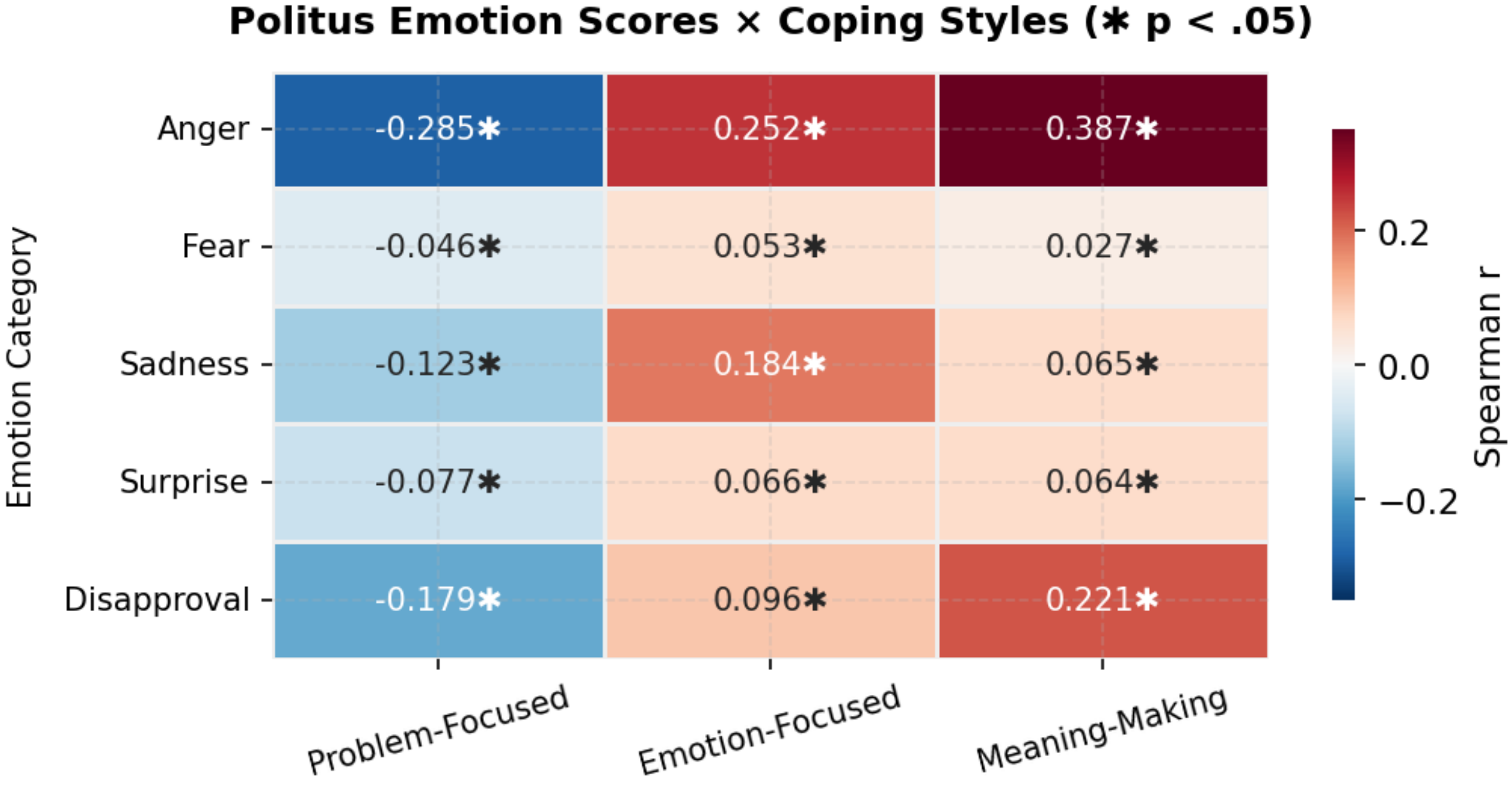

*Figure 5. Heatmap of Spearman correlations between Politus emotion categories and predicted coping styles. * p <.05.*

Anger shows the largest correlations of any emotion: r = −0.285 with problem-focused and r = 0.387 with meaning-making — the most substantively meaningful associations in this analysis. This pattern suggests that angry users are less likely to engage in practical aid coordination and more likely to assign moral or political meaning to the disaster — a finding consistent with Jin's (2009) cognitive appraisal model, which links anger to other-blame and accountability demands. Sadness correlates most strongly with emotion-focused coping (r = 0.184), consistent with its role as a grief processing strategy.

### 6. Discussion

The findings of this study show that when millions of people face the same disaster at the same time, their collective response is not chaotic. In contrast, it follows recognizable psychological patterns that move in predictable ways over time. Three contributions follow from this observation.

First, coping styles can be reliably detected at tweet level using a fine tuned Turkish BERT model. The macro F1 of 0.693 — achieved with only 500 annotated examples suggests that pre-trained language models carry substantial transferable knowledge about coping-relevant language patterns, even in a low-resource setting. The dramatic performance gap between BERTurk and the zero-shot mDeBERTa baseline (macro F1 0.693 vs. 0.324) reflects the importance of language-specific models for Turkish NLP tasks. This result also has a practical implication: low-resource NLP is viable for psychological classification in non-English crisis contexts. It opens the door for similar studies in languages and communities that have been historically underrepresented in computational social science.

The temporal findings offer what is the first large-scale empirical test of coping phase shifts in digital crisis discourse. Problem-focused coping dominated the first three days, followed by emotion-focused processing in the middle weeks, and meaning-making grew steadily throughout. It is a sequence consistent with both coping theory (Lazarus & Folkman, 1984) and crisis communication research (Coombs, 2015). The strength of the meaning-making trajectory, however, is harder to explain from theory alone. By Phase 4, meaning-making reached 0.373, which is a level that likely reflects not only general coping dynamics but also the specific context of this disaster. The earthquake struck three months before a national election. Public anger had a clear institutional target. The building collapses were widely linked to weak construction enforcement. Under these conditions, the turn toward blame attribution and political meaning-making was strong. Coping dynamics, these findings suggest, are shaped by the political landscape in which disasters unfold.

Also, the emotion–coping correlation analysis reveals a nuanced mapping between discrete emotions and coping strategies. The strong association between anger and meaning making is particularly theoretically significant. It suggests that anger channels into interpretive blame attribution rather than practical problem-solving, a finding with implications for understanding the political dynamics of disaster response. This pattern is consistent with Jin's (2009) cognitive appraisal model, which links anger specifically to other-blame and accountability demands. In the context of the 2023 earthquake, anger directed public attention toward institutional failure and political responsibility. Whether this translated into political action, however, is a question the present data cannot answer.

## 7. Limitations

The avoidance coping category could not be reliably operationalized ($\kappa$ = 0.021), suggesting this strategy either manifests differently in short-form social media text or requires a substantially different annotation approach. The annotation training set (n = 500) is small, and the moderate inter-annotator agreement for meaning-making ($\kappa$ = 0.449) indicates this category remains difficult to operationalize consistently. Future work integrating multimodal data and larger annotated corpora may address these limitations.

Beyond methodological constraints, the study also faces limitations of representativeness. The earthquake most severely affected Kahramanmaraş and Hatay, yet social media discourse is not a representative sample of those affected. Rural communities and older populations, who bore a disproportionate share of the loss, may cope through channels this study cannot observe: face-to-face networks, religious communities, or silence. Future work combining social media data with survey or ethnographic methods could help triangulate these findings against less visible forms of coping.

A further limitation concerns the interpretation of coping style distributions, particularly in Phase 1. The high proportion of problem-focused tweets in the immediate aftermath of the earthquake may not solely reflect individuals' psychological coping dispositions. During an active emergency, writing 'reach AFAD at this number' or sharing a trapped person's location is a situation requirement as much as it is a coping choice. The environment itself invites and rewards this type of behaviour This raises a distinction between coping style as a psychological tendency and textual demand as a situational constraint. While this study acknowledges in the introduction that tweets are public textual traces rather than transparent windows into psychology, the specific entanglement of situation and disposition in Phase 1 is worth flagging explicitly. Future work combining social media data with survey or experience sampling methods could help distinguish these two sources of variation.

## 8. Conclusion

This study presents the first large-scale computational operationalization of psychological coping theory in Turkish-language digital crisis discourse. Using over one million tweets from the 2023 Türkiye earthquake, we train and validate a multi-label BERTurk classifier and apply it to track coping style dynamics across four post-disaster phases. Our findings confirm theoretical predictions about the temporal sequencing of problem-focused, emotion-focused, and meaning making coping, while revealing novel patterns in the relationship between anger, disapproval, and meaning attribution.

The results demonstrate both the feasibility and the value of bridging coping psychology and computational NLP. But beyond methodology, this study points toward a practical question: if we can detect how a population is coping in real time, what can we do with that knowledge?

Coping styles are not just descriptive categories, they signal different needs. A population dominated by problem-focused coping needs information, coordination, and access to resources. A population in the emotion-focused phase needs acknowledgment, presence, and space to grieve. When meaning-making peaks, people are asking harder questions about responsibility, about institutional failure, about what kind of society allows this to happen. Each phase calls for a different response.

Understanding this trajectory can help humanitarian organizations and crisis communicators design their interventions to where a community actually is, rather than where responders assume it to be. For instance, when emotion-focused coping dominates, crisis communicators might prioritize creating spaces for collective mourning, coordinated memorial campaigns, community grief support channels, or public acknowledgment of loss by institutional actors (cf. Coombs, 2015). When meaning-making rises, transparent institutional communication and visible accountability mechanisms become more important than logistical updates.

This study offers a methodological foundation for building such real-time coping dashboards. At the same time, any deployment of psychological classification tools in crisis contexts must be approached with careful attention to privacy, consent, and the risk of misuse, concerns that apply with particular force when the affected population is already vulnerable (Hovy & Spruit, 2016). The goal is not surveillance but understanding to read what people are going through in the moments they did not anticipate, and to respond in kind.